\documentclass[conf]{IEEEtran}
\usepackage{multirow}
\usepackage{algorithm,algpseudocode}
\usepackage{amsmath}
\usepackage{amssymb}
\usepackage{mathtools}

\usepackage{cite}
\def\BibTeX{{\rm B\kern-.05em{\sc i\kern-.025em b}\kern-.08em
    T\kern-.1667em\lower.7ex\hbox{E}\kern-.125emX}}
\usepackage[bookmarksnumbered=true]{hyperref} 
\hypersetup{
     colorlinks = true,
     linkcolor = blue,
     anchorcolor = blue,
     citecolor = blue,
     filecolor = blue,
     urlcolor = blue
     }

\usepackage{svg}

\usepackage{wrapfig}
\usepackage{caption}
\usepackage{subcaption}
\usepackage{booktabs}

\usepackage{tikz}
\def\checkmark{\tikz\fill[scale=0.4](0,.35) -- (.25,0) -- (1,.7) -- (.25,.15) -- cycle;} 

\newcommand{\Lagr}{\mathcal{L}}
\usepackage{xcolor}
\begin{document}

\title{MIME: Adapting a Single Neural Network for \underline{M}ulti-task \underline{I}nference with \underline{M}emory-\underline{e}fficient Dynamic Pruning}

\author{
Abhiroop Bhattacharjee, Yeshwanth Venkatesha, Abhishek Moitra, and Priyadarshini Panda \\
\{abhiroop.bhattacharjee, yeshwanth.venkatesha, abhishek.moitra, priya.panda\}@yale.edu \\

Department of Electrical Engineering, Yale University, USA

}

\maketitle

\begin{abstract}

Recent years have seen a paradigm shift towards multi-task learning. This calls for memory and energy-efficient solutions for inference in a multi-task scenario. We propose an algorithm-hardware co-design approach called MIME. MIME reuses the weight parameters of a trained parent task and learns task-specific threshold parameters for inference on multiple child tasks. We find that MIME results in highly memory-efficient DRAM storage of neural-network parameters for multiple tasks compared to conventional multi-task inference. In addition, MIME results in input-dependent dynamic neuronal pruning, thereby enabling energy-efficient inference with higher throughput on a systolic-array hardware. Our experiments with benchmark datasets (child tasks)- CIFAR10, CIFAR100, and Fashion-MNIST, show that MIME achieves $\sim3.48\times$ memory-efficiency and $\sim2.4-3.1\times$ energy-savings compared to conventional multi-task inference in \textit{Pipelined task mode}.

\end{abstract}


\IEEEpeerreviewmaketitle

\section{Introduction}
\label{sec:intro}

Deep Neural Networks (DNNs) have increasingly been deployed for various applications ranging from computer vision, voice recognition to natural language processing and so forth \cite{krizhevsky2012imagenet, hinton2012deep, goldberg2016primer}. Furthermore, in today's era of \textit{Internet-of-Things}, many of these applications need to operate in highly resource-constrained environments. As a result, designing efficient hardware accelerators for memory and energy-efficient implementation of DNNs has become imperative \cite{peng2019dnn+, chen2016eyeriss, bhattacharjee2021neat, biswas2018conv}. 


\begin{wrapfigure}{l}{0.2\textwidth}
\includegraphics[width=0.2\textwidth]{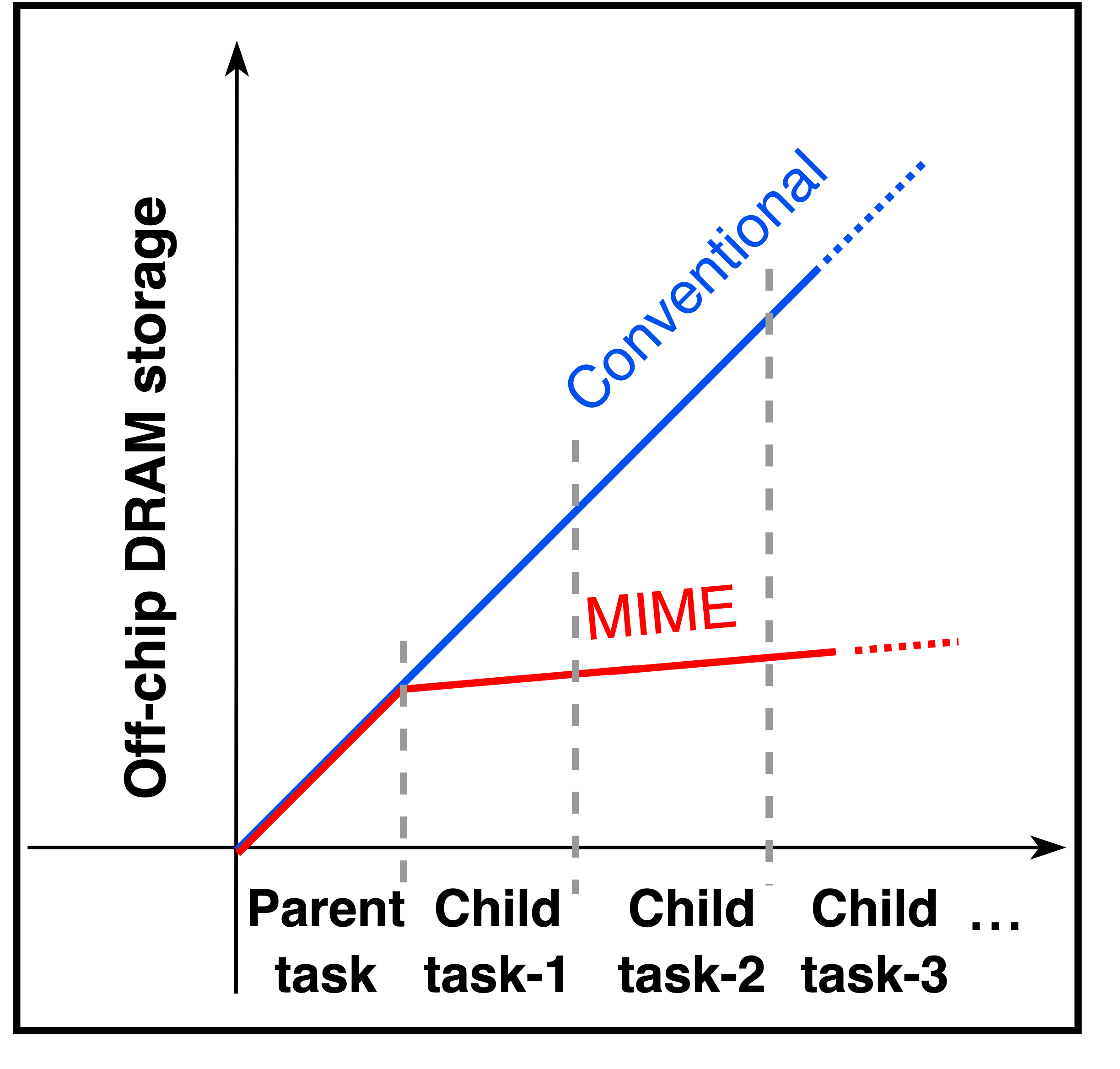}
  
\caption{A representation of the off-chip DRAM storage for conventional multi-task inference scenario (blue) and MIME (red)}
\label{dram_mime}
\end{wrapfigure}

Data such as images can have several common features that can be leveraged for multiple classification tasks. Thus, recent years have witnessed efforts to shift towards the multi-task learning paradigm \cite{pan2009survey, oquab2014learning, lu2021convolutional, fu2021learn}. To achieve multi-task learning, several algorithms have been proposed that use transfer learning. Traditional transfer learning techniques train a DNN model for a parent task (or dataset) and then fine-tune its parameters for multiple downstream tasks (or datasets) called child tasks \cite{tan2018survey}. This reduces the complexity of training a different neural network from scratch for every task. 

Conventional task-specific fine-tuning during multi-task learning assumes the following- (i) every child task generates a new set of weight parameters. During inference on hardware, all of these parameters need to be stored in the off-chip DRAM that pose huge memory overhead (see blue curve in Fig. \ref{dram_mime}). (ii) for multi-task inference, a batch of inputs fed into a DNN on hardware consists of data (images) that belong to a single task (referred to as \textit{Singular task mode}) \cite{whatmough2019fixynn, nurvitadhi2016hardware, nurvitadhi2015sparse}. However, let us consider a more realistic scenario wherein, a batch of inputs can have images belonging to multiple tasks in an interleaved fashion (referred to as \textit{Pipelined task mode}). In this case, the number of accesses to the DRAM for fetching the task-specific weight parameters would increase significantly leading to high energy overhead. 


\begin{figure}[t]
    \centering
    \includegraphics[width=\linewidth]{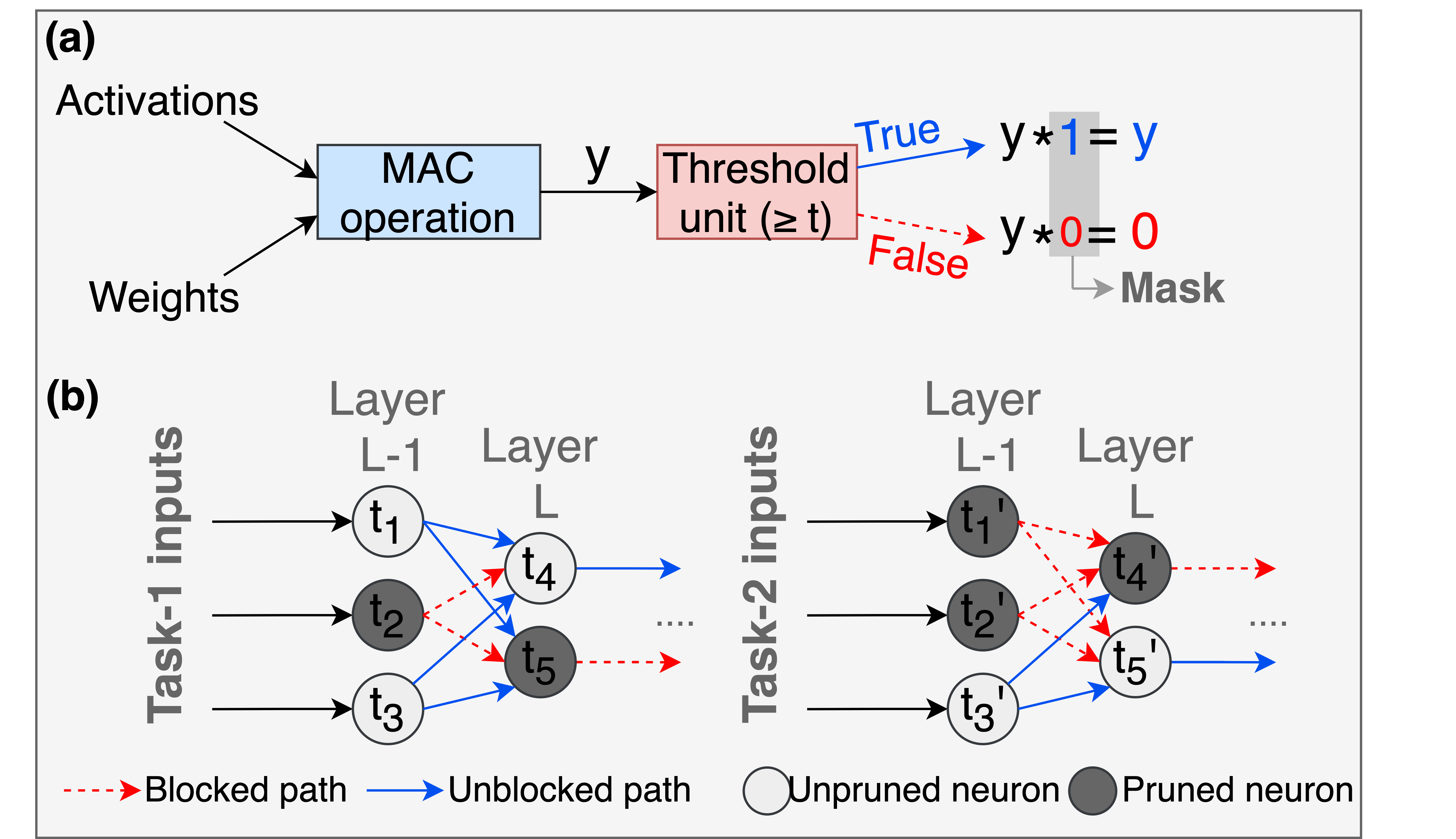}
   
    \caption{(a) Pictorial representation of the flow during forward propagation in MIME. It shows threshold-induced binary mask generation; (b) A representation of task and input-dependent dynamic neuronal pruning due to MIME. The neuronal threshold parameters are represented as $t_i$ and $t_{i}^{'}$ for task-1 and task-2 inputs, respectively.}
    \label{mem_sav}
     \vspace{-7mm}
\end{figure}

To circumvent the above challenges, we propose a technique for memory and energy-efficient multi-task inference on hardware. Our algorithm-hardware co-simulation approach is called MIME. In MIME, instead of fine-tuning the weights of the parent model for multiple child tasks during transfer learning, we propose an algorithm to learn task-specific threshold parameters for multi-task inference. Here, the weight parameters of the parent task ($W_{parent}$) are re-utilized across all the child tasks. As can be seen in Fig. \ref{mem_sav}(a) \& (b), each neuron in a DNN is associated with a task-specific threshold parameter ($t$) against which the Multiply-and-Accumulate (MAC) output ($y$) of the neuron is compared and a binary mask (0 or 1) is generated. If the mask has a value equal to 1, the corresponding neuron is active and produces $y$ as its output activation. Otherwise, the neuron is pruned and produces a zero activation value. Thus, based on the input and the child task being processed by the DNN, the corresponding set of threshold parameters are selected and different sub-networks within the same DNN model are activated during inference (shown in Fig. \ref{mem_sav}(b)). As we will see in the upcoming sections, this input and task-dependent dynamic neuronal pruning manifests as reduction in computational and communication energies as well as increased throughput on hardware \cite{yang2017designing}. 

MIME requires the storage of threshold parameters specific to every child task in conjunction with $W_{parent}$ in the DRAM for inference. This is in contrast to storing a new set of fine-tuned weight parameters for every child task during conventional multi-task scenario. In this work, we show that the DRAM storage for the weight and threshold parameters across all the child tasks in case of MIME is significantly lower than the storage of weight parameters corresponding to each child task for conventional multi-task inference (see red curve in Fig. \ref{dram_mime}). This makes MIME highly memory-efficient. 

In summary, the key contributions of this work are as follows:

\begin{itemize}
    \item This work, for the first time, raises a practical/realistic question for multi-task scenarios, that is, \textit{How to improve energy-efficiency during inference when inputs in a batch belong to different tasks (i.e. Pipelined task mode)?} We propose MIME to enable energy and memory-efficient multi-task DNN inference on hardware. In MIME, the weight parameters of a parent task $W_{parent}$ are reused during the inference of multiple child tasks. 
    
    \item We propose an algorithm to learn the threshold parameters corresponding to the child tasks ($T_{child}$)  that are used in conjunction with $W_{parent}$ for inference without huge training overhead. 
    
    
    \item We evaluate the performance of MIME on \textit{Eyeriss}-systolic-array based hardware  architecture (in 65 nm CMOS technology) \cite{chen2016eyeriss, yang2018energy} under two modes, namely \textit{Singular task mode} and \textit{Pipelined task mode}. For both the modes of operation, we adopt an \textit{output stationary} (OS) dataflow for inference. We conduct ablation studies to suggest important design metrics for selecting the best trade-off between compute energy-and-memory cost.
    
    \item We perform comprehensive experiments using a VGG16 DNN~\cite{vgg} on benchmark datasets---Imagenet \cite{krizhevsky2012imagenet} as parent task and, CIFAR10, CIFAR100 \cite{cifar} and Fashion-MNIST \cite{xiao2017fashion} as child tasks. We find that in the \textit{Pipelined task mode}, MIME leads to significantly lower DRAM accesses compared to the conventional multi-task inference scenario, thereby bringing in huge layerwise energy savings ($\sim2.4-3.1\times$). MIME also leads to a significant layerwise improvement in throughput during inference ($\sim2.8-3.0\times$), primarily attributed to the dynamic neuronal sparsity.
    

\end{itemize}

\section{Related works}

\begin{table*}[t]
\centering
\caption{Table showing comparison with related works}
\label{tab:comparison}
\begin{tabular}{|c|cccc|}
\hline
\textbf{}                   & \multicolumn{4}{c|}{\textbf{Objectives of the works}}                                                                \\ \hline
\multirow{2}{*}{\textbf{Works}} &
  \multicolumn{1}{c|}{\multirow{2}{*}{\textbf{Energy \& Memory efficiency}}} &
  \multicolumn{2}{c|}{\textbf{Multi-task learning}} &
  \multirow{2}{*}{\textbf{Training complexity reduction}} \\ \cline{3-4}
                            & \multicolumn{1}{c|}{}   & \multicolumn{1}{c|}{\textbf{Sequential}} & \multicolumn{1}{c|}{\textbf{Simultaneous}} &    \\ \hline
\textbf{Transfer learning}  & \multicolumn{1}{c|}{--} & \multicolumn{1}{c|}{--}                  & \multicolumn{1}{c|}{\textbf{\checkmark}}                      &   \textbf{\checkmark} \\ \hline
\textbf{Pruning}            & \multicolumn{1}{c|}{\textbf{\checkmark}}   & \multicolumn{1}{c|}{--}                  & \multicolumn{1}{c|}{--}                    & -- \\ \hline
\textbf{Continual learning} & \multicolumn{1}{c|}{--} & \multicolumn{1}{c|}{\textbf{\checkmark}}                    & \multicolumn{1}{c|}{--}                    & -- \\ \hline
\textbf{Our work (MIME)}    & \multicolumn{1}{c|}{\textbf{\checkmark}}   & \multicolumn{1}{c|}{--}                  & \multicolumn{1}{c|}{\textbf{\checkmark}}                      &   \textbf{\checkmark} \\ \hline
\end{tabular}
\end{table*}

\textbf{Conventional transfer learning: }There has been a body of work on transfer learning wherein the weight parameters of a DNN model trained for a parent task are fine-tuned to run a downstream child task \cite{tan2018survey}. On similar lines, knowledge distillation works enable learning of smaller child models by distilling the loss function of the larger parent model \cite{hinton2015distilling, gou2021knowledge}. Such approaches have been shown to significantly reduce the training complexity of the child models. However, when there are several downstream tasks with each task having its own set of weight parameters, traditional transfer learning or distillation approaches do not provide memory and energy-efficient solutions to store and access the parameters during inference on hardware. 

\textbf{Pruning strategies:}  Several pruning techniques have been devised to generate highly compressed and sparse DNNs. They can be categorized as static (only weights are pruned) or dynamic (both weights and activations are pruned). The sparse DNNs when deployed on hardware lead to high memory and energy-efficiencies during inference \cite{yang2017designing,liang2021pruning, you2020rsnn, wang2020high, sun2020computation, chu2020pim, zhang2021hardware}. However, all prior pruning approaches only cater to a single task scenario with the pruning strategy defined for the given dataset/model. 

\textbf{Continual learning: } There have been recent works on multi-task continual learning wherein data from numerous tasks (or numerous segments of a task) are \textit{sequentially} shown to learn a DNN model \cite{parisi2019continual, ramesh2021boosting}. In contrast, MIME works under the assumption that the entire data for a child task is available. MIME \textit{simultaneously} learns task-specific threshold parameters of the parent DNN model for multiple downstream child tasks keeping the parent weights frozen.  

Table \ref{tab:comparison} provides a qualitative comparison between MIME and the above related works highlighting our key contributions.

\vspace{-2mm}

\section{Methodology and System Implementation}

\subsection{Task-specific threshold generation for MIME}
\label{sec: th_mime}

\begin{figure*}[t]
    \centering
    \includegraphics[width=\linewidth]{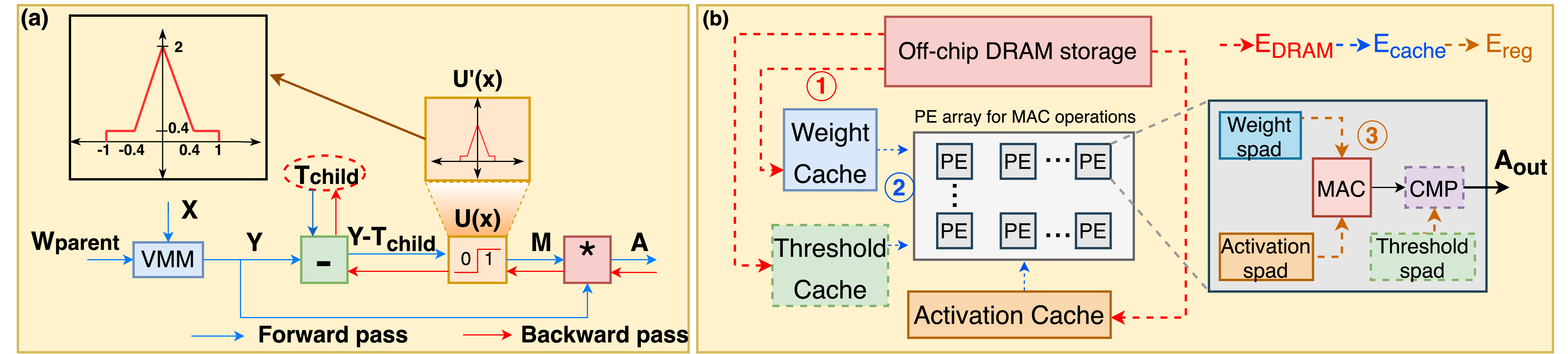}

    \caption{(a) Flow diagram showing the forward (blue) and backward (red) propagation steps involved in training task-specific threshold parameters for MIME. Note, $Y$, $M$ and $A$ respectively denote the entire matrix of MAC outputs ($y_i$s), binary masks ($m_i$s) and output activations ($a_i$s) for $i^{th}$ neuron; (b) Pictorial representation of the dataflow in a systolic-array hardware architecture during inference. The memory hierarchy includes: (1) accessing data from the off-chip DRAM and storing into the cache, (2) fetching task-specific parameters and activations from the cache to the spads, and (3) performing MAC operations in the PEs by fetching operands from the spads.}
    \label{method_flow}
     \vspace{-5mm}
\end{figure*}

Consider a DNN model trained for a parent task with its weight parameters denoted by $W_{parent}$. Our objective is to carry out inference using $W_{parent}$ across multiple downstream tasks or child tasks. We, thus, propose an algorithm to train a set of threshold parameters specific to a child task ($T_{child}$), used in conjunction with $W_{parent}$.
Training the threshold parameters with the child datasets includes forward and backward propagation through the DNN model as described in Fig. \ref{method_flow}(a). During the forward pass, after getting the \textit{Vector-Matrix-Multiplications} (VMMs) with DNN weights ($W_{parent}$) and input activations ($X$), we compare the VMM value of the $i^{th}$ output neuron ($y_{i}$) with a threshold parameter ($t_{i}>0$) to generate a mask ($m_{i}$) as follows:
\begin{equation}
    m_{i} =
\begin{cases}
 1 ,   & y_{i} - t_{i} \geq  0 \\
    0 , & y_{i} - t_{i} < 0
\end{cases}
\label{eq:mask}
\end{equation}

The final output activation of the $i^{th}$ neuron ($a_{i}$) is computed via masking the VMM outputs ($y_{i}$) as follows:

\begin{equation}a_{i} = y_{i} * m_{i}
\label{eq:out}
\end{equation}

During backward propagation, we keep all the weights $W_{parent}$ frozen and only update the threshold $T_{child} = \{t_1, t_2....t_i\}$, where $t_i$ is associated with each neuron $i$ in the DNN. To circumvent the non-differentiable nature of the mask-generation step function (equation \ref{eq:mask}), we estimate the gradient using a piece-wise linear polynomial function as shown in Fig. \ref{method_flow}(a) \cite{liu2020dynamic}.
The loss function ($\Lagr$) during the training of the $t_i$ parameters is defined as follows:

\begin{equation}\Lagr = \Lagr_{CE} + \beta*\Lagr_{t}
\label{eq:loss_func}
\end{equation}

where, $\beta$ is a hyper-parameter that assumes a value of $1e-6$ for training with a batch size of 100, $\Lagr_{CE}$ denotes the cross-entropy loss and the \textit{threshold-regularization term} ($\Lagr_{t}$) is defined as:
\begin{equation}\Lagr_{t} = \sum_{\forall layers}^{}\sum_{\forall i}^{}{exp(t_{i})}
\label{eq:Lt}
\end{equation}

The \textit{threshold-regularization term} ($\Lagr_{t}$) prevents the threshold parameters ($t_{i}$) from assuming arbitrarily large positive values, which would otherwise result in convergence issues. 
At the end of the training for $n$ child tasks, MIME yields the parameters:\{$W_{parent}$, $T_{child-1}$, $T_{child-2}$,..., $T_{child-n}$\} that need to be stored for inference on hardware.
It is evident that the masking due to thresholds will yield input-dependent dynamic sparsity at every DNN layer that translates to energy savings and high throughput during hardware implementation. 

\subsection{Implementation of MIME on a systolic-array hardware}
\label{sec: hwd_mime}

To understand the implications of generating task-specific thresholds for carrying out inference on a systolic-array hardware architecture \cite{chen2016eyeriss, yang2018energy}, we consider two modes of inference: \textit{Singular task mode} and \textit{Pipelined task mode}. 

As shown in Fig. \ref{method_flow}(b), for a given DNN layer during inference, first the corresponding weight (of the parent task) and task-specific threshold parameters are respectively loaded into the weight and threshold caches and the previous layer's non-zero activations are loaded into the activation cache from the off-chip DRAM. For the \textit{Pipelined task mode} with MIME, even if the subsequent non-zero activations in the queue belongs to a different task (dataset), the weight parameters of the given DNN layer need not be reloaded. In contrast, in the conventional multi-task scenario wherein, each task has its own set of weight parameters, there has to be multiple fetch-load cycles of weight parameters. In MIME, pertaining to the task, only the threshold parameters need to be reloaded into the threshold cache from the DRAM, which typically has a lesser overhead than reloading weight parameters to the weight cache. This translates to higher energy-savings as we will see in the upcoming sections.

Next, to perform the VMM or MAC operations in the PEs, the operands are fetched from the caches to the scratchpads (spads) or local registers situated inside the PEs. Here, the fetches from the cache are carried out only for those weights which interact with the non-zero activations for MAC operations. Thus, the layerwise neuronal sparsity arising due to threshold-induced dynamic masking results in both compute energy and memory access savings due to zero-skipping. 
Also, we follow an OS dataflow in carrying out MAC operations in the PE array. Since, each output neuron of a convolutional layer is associated with a threshold parameter, OS dataflow helps reduce repeated accesses of the threshold parameters as well as the partial sums to and from the main memory. Inside the PEs, there are MAC computation and comparator (CMP) units that fetch operands from the local spads and compute the final masked output neuronal activations that are stored back into the off-chip DRAM.

\vspace{-4mm}
\section{Experiments}
\label{sec: expt}

\begin{table*}[t]
\centering
\caption{Table showing test accuracy and average layerwise neuronal sparsity for VGG16 DNN for the child datasets (CIFAR10, CIFAR100 and F-MNIST) using MIME}
\label{tab:mime_acc}
\resizebox{\linewidth}{!}{%
\begin{tabular}{@{}ccccccccccccc@{}}
\toprule
\multicolumn{1}{l}{}   & \multicolumn{1}{l}{} & \multicolumn{11}{c}{\textbf{Average layerwise   neuronal sparsity (due to MIME)}}                             \\ \midrule
\multicolumn{1}{c}{\textbf{Child task}} &
  \multicolumn{1}{c}{\textbf{Test Accuracy (\%)}} &
  \multicolumn{1}{c}{\textbf{conv2}} &
  \multicolumn{1}{c}{\textbf{conv4}} &
  \multicolumn{1}{c}{\textbf{conv5}} &
  \multicolumn{1}{c}{\textbf{conv7}} &
  \multicolumn{1}{c}{\textbf{conv8}} &
  \multicolumn{1}{c}{\textbf{conv9}} &
  \multicolumn{1}{c}{\textbf{conv10}} &
  \multicolumn{1}{c}{\textbf{conv12}} &
  \multicolumn{1}{c}{\textbf{conv13}} &
  \multicolumn{1}{c}{\textbf{conv14}} &
  \multicolumn{1}{c}{\textbf{conv15}} \\ \midrule
\textbf{CIFAR10}       & 83.57                  & 0.6493 & 0.6081 & 0.6587 & 0.6203 & 0.6233 & 0.6449 & 0.6679 & 0.6477 & 0.6553 & 0.6855 & 0.657 \\
\textbf{CIFAR100}      & 59.42                  & 0.6522 & 0.5951 & 0.6373 & 0.6100 & 0.6121 & 0.6279 & 0.6580 & 0.6374 & 0.6388 & 0.6703 & 0.6571 \\
\textbf{F-MNIST} & 88.36                  & 0.6075 & 0.5634 & 0.6138 & 0.5991 & 0.5959 & 0.6017 & 0.6204 & 0.6014 & 0.6125 & 0.6138 & 0.6287 \\ \bottomrule
\end{tabular}%
}
\end{table*}

\begin{table*}[t]
\centering
\caption{Table showing test accuracy and average layerwise neuronal sparsity for VGG16 DNN for the baseline models (CIFAR10, CIFAR100 and F-MNIST) using conventional multi-task inference}
\label{tab:base_acc}
\resizebox{\linewidth}{!}{%
\begin{tabular}{@{}ccccccccccccc@{}}
\toprule
\multicolumn{1}{l}{}   & \multicolumn{1}{l}{} & \multicolumn{11}{c}{\textbf{Average layerwise   neuronal sparsity (due to ReLU)}}                             \\ \midrule
\multicolumn{1}{c}{\textbf{Baseline Child task}} &
  \multicolumn{1}{c}{\textbf{Test Accuracy (\%)}} &
  \multicolumn{1}{c}{\textbf{conv2}} &
  \multicolumn{1}{c}{\textbf{conv4}} &
  \multicolumn{1}{c}{\textbf{conv5}} &
  \multicolumn{1}{c}{\textbf{conv7}} &
  \multicolumn{1}{c}{\textbf{conv8}} &
  \multicolumn{1}{c}{\textbf{conv9}} &
  \multicolumn{1}{c}{\textbf{conv10}} &
  \multicolumn{1}{c}{\textbf{conv12}} &
  \multicolumn{1}{c}{\textbf{conv13}} &
  \multicolumn{1}{c}{\textbf{conv14}} &
  \multicolumn{1}{c}{\textbf{conv15}} \\ \midrule
\textbf{CIFAR10}       & 84.25                  & 0.4983 & 0.4506 & 0.5390 & 0.5015 & 0.5097 & 0.5341 & 0.5635 & 0.5358 & 0.5420 & 0.5627 & 0.5608 \\
\textbf{CIFAR100}      & 60.55               & 0.5030 & 0.4586 & 0.5399 & 0.5069 & 0.5129 & 0.5333 & 0.5633 & 0.5345 & 0.5449 & 0.5842 & 0.6002 \\
\textbf{F-MNIST} & 90.12                  & 0.5114 & 0.4796 & 0.5488 & 0.5230 & 0.5260 & 0.5329 & 0.5503 & 0.5280 & 0.5343 & 0.5507 & 0.5820 \\ \bottomrule
\end{tabular}%
}
\vspace{-4mm}
\end{table*}

We take a trained VGG16 DNN with Imagenet dataset (parent dataset) with $73.36\%$ test accuracy, and obtain the $W_{parent}$ parameters. Next, using $W_{parent}$, we train the VGG16 DNN for the child tasks (datasets), namely CIFAR10, CIFAR100 and Fashion-MINST (F-MNIST) to obtain $T_{child-1}$, $T_{child-2}$ and $T_{child-3}$, respectively. Note, CIFAR10 and CIFAR100 are two similar datasets or tasks consisting of RGB images of size $32\times32$, while, the F-MNIST dataset consists of grayscale images of size $28\times28$. We considered such different types of datasets to show that our method is transferable from one parent task to different kinds of child tasks. Training for the task-specific threshold parameters was carried out using the methodology shown in Fig. \ref{method_flow} for 10 epochs using ADAM optimizer with a learning rate of $1e-3$. Thus, MIME incurs very low training overhead. The DNN test accuracies for the child tasks have been reported in Table \ref{tab:mime_acc}. We also present the average layerwise sparsity in the output activations, observed with MIME, for the VGG16 DNN across different child tasks in Table \ref{tab:mime_acc}. 

To analyze the benefits of MIME on hardware, we present the test accuracies and layerwise neuronal sparsities for our baseline models in Table \ref{tab:base_acc}. Note, the baselines are generated by normally training the VGG16 DNN on three child datasets and obtaining $W_{child-1}$, $W_{child-2}$ and $W_{child-3}$ weight-parameters for each. Here, the average sparsity in the output activations for each layer arises due to the ReLU operation that masks out the negative MAC outputs for each neuron. 

Next, we implement the above models on the systolic-array architecture. For inference in \textit{Singular task mode}, we consider a batch consisting of three input images, each belonging to one task (say, CIFAR10) and present our hardware analyses for this batch of inputs (see Section \ref{sec:single-task}). For \textit{Pipelined task mode}, we again consider a batch of three input images in succession belonging to three different tasks or datasets - CIFAR10, CIFAR100 and F-MNIST (see Section \ref{sec:pip-task}). In this study, we assume that the hardware has knowledge about the task it is currently processing and thus, can accordingly fetch the right set of parameters from the memory to the MAC compute units for inference. Unless otherwise stated, the specifications pertaining to the systolic-array  accelerator are values listed in Table \ref{tab:eyeriss}. Note, all energy values have been normalized with respect to the absolute energy of 1 MAC operation in the PE. 

\begin{table}[t]
\centering
\caption{Table showing system parameters for the systolic-array hardware. Here, $e_{DRAM}$, $e_{cache}$ and $e_{reg}$ are respectively the energies corresponding to 1 DRAM, 1 cache and 1 spad based memory-access normalized \textit{w.r.t.} energy of 1 MAC operation ($e_{MAC}$) \cite{chen2016eyeriss}}
\label{tab:eyeriss}
\begin{tabular}{|c|c|}
\hline
\textbf{Parameter} & \textbf{Value} \\ \hline
Technology     & 65 nm CMOS        \\ \hline
Precision ($W,X,A,T$)     & 16 bits        \\ \hline
\begin{tabular}[c]{@{}c@{}}Cache size\\ (Activation, Weight, Threshold)\end{tabular} & 156 KB \\ \hline
Spad size          & 512 B          \\ \hline
PE array size      & 1024           \\ \hline
$e_{DRAM}$              & $200\times$           \\ \hline
$e_{cache}$             & $6\times$             \\ \hline
$e_{reg}$               & $2\times$             \\ \hline
$e_{MAC}$               & $1\times$             \\ \hline
\end{tabular}%
\vspace{-4mm}
\end{table}

\vspace{-4mm}

\section{Results and Discussion}
\label{sec: results}

\subsection{Reduction in off-chip DRAM storage}
\label{sec:dram}

  

\begin{figure}[t]
    \centering
    \includegraphics[width=0.7\linewidth]{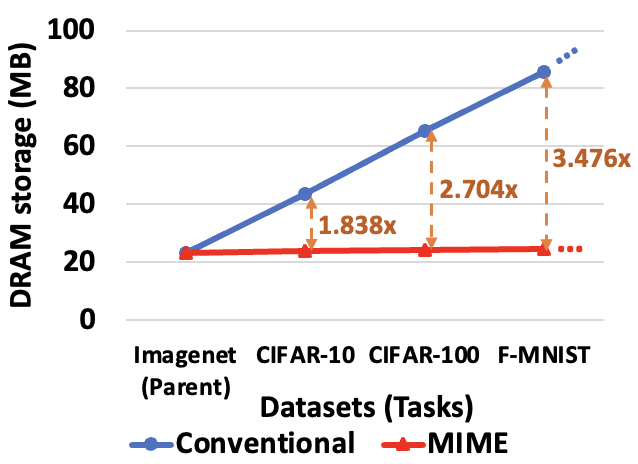}
    
    \caption{Plot showing savings in off-chip DRAM storage due to MIME (red) compared with conventional multi-task inference scenario (blue)}
    \label{dram_mem}
     \vspace{-6mm}
\end{figure}

Fig. \ref{dram_mem} presents the total off-chip DRAM storage needed for storing the weights and threshold parameters of the parent task (Imagenet) and its child tasks (CIFAR10, CIFAR100 and F-MNIST) for MIME. As discussed in Section \ref{sec:intro}, we find that for the parent task and its $n$ downstream child tasks, the memory savings with respect to conventional multi-task inference approach is $>n\times$ as has been annotated in the figure. For the Imagenet task along with CIFAR10, CIFAR100 and F-MNIST tasks (3 child tasks), we obtain $\sim 3.48\times$ savings in DRAM storage. The absolute value of the savings in off-chip DRAM storage with MIME increases further with increase in the number of child task for the given parent task. This makes our approach highly memory-efficient. 

\vspace{-4mm}

\subsection{Results for Singular Task Mode}
\label{sec:single-task}

\begin{figure}[t]
    \centering
    \includegraphics[width=0.8\linewidth]{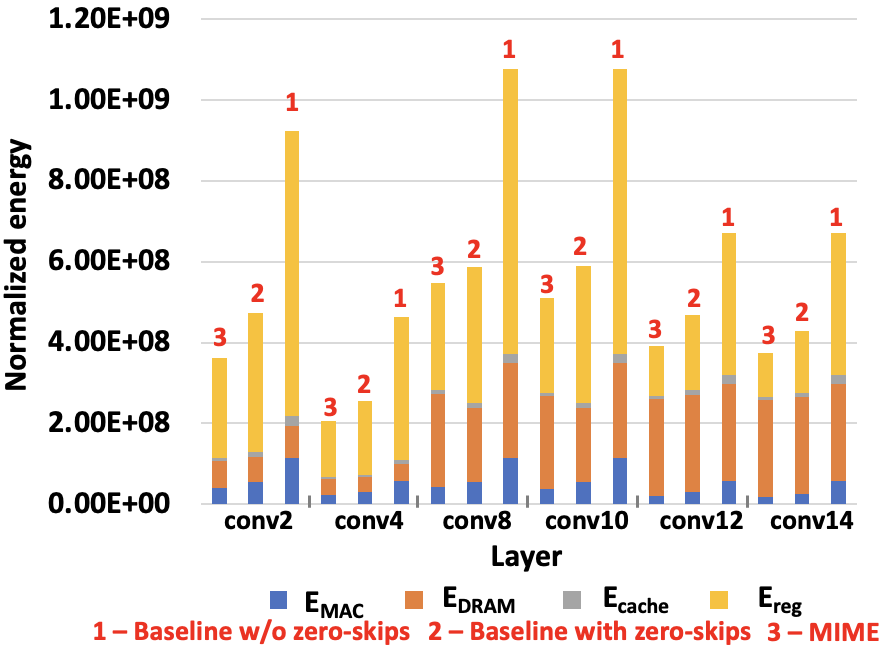}
    
    \caption{Plot showing layerwise energy distribution for the convolutional layers of VGG16 DNN implemented on systolic-array hardware for multi-task inference in \textit{Singular task mode}}
    \label{energy_sing}
     \vspace{-4mm}
\end{figure}

In Fig. \ref{energy_sing}, we plot the energy distribution (normalized \textit{w.r.t.} energy of 1 MAC operation) of the convolutional layers of the VGG16 DNN processing a batch of 3 input images from the CIFAR10 dataset. The overall layerwise energy is distributed among total energy due to DRAM accesses ($E_{DRAM}$), cache accesses ($E_{cache}$), scratchpad accesses in the PEs ($E_{reg}$) and the total energy expended during MAC computations ($E_{MAC}$). The results are presented for three cases- \textbf{Case-1:} Using Baseline task-models (in Table \ref{tab:base_acc}) without skipping computations and communications for zero-valued activations, \textbf{Case-2:} Using Baseline task-models (in Table \ref{tab:base_acc}) and skipping computations and communications for zero-valued activations, and \textbf{Case-3:} Using the MIME approach. Note, for brevity, we show results pertaining to the even-numbered convolutional layers in the VGG16 DNN. 

In the \textit{Singular task mode}, the energy savings in case of MIME with respect to Case-1 or Case-2 are primarily attributed to the dynamic neuronal pruning at each layer. Quantitatively, we obtain $\sim1.8-2.5\times$ energy savings with MIME with respect to baseline Case-1 and $\sim1.07-1.30\times$ with respect to baseline Case-2. However, it is clear from Fig. \ref{energy_sing} that the benefits of reduced DRAM accesses and hence, lower values of $E_{DRAM}$, cannot be seen for the \textit{Singular task mode} scenario with MIME. In fact, $E_{DRAM}$ of MIME is slightly higher than the corresponding $E_{DRAM}$ of Case-2 for each layer. This is because in addition to weight parameters, the threshold parameters also need to be fetched from the DRAM for MAC operations. Thus, to reap the benefits of MIME approach, we consider the \textit{Pipelined task mode} of inference in the next section. 

\vspace{-4mm}

\subsection{Results for Pipelined Task Mode}
\label{sec:pip-task}

\begin{figure}[t]
    \centering
    \includegraphics[width=0.8\linewidth]{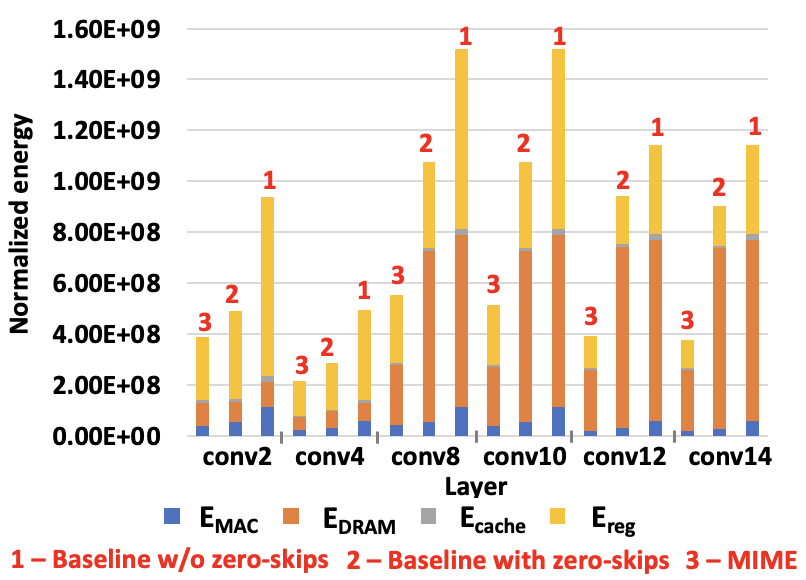}
    
    \caption{Plot showing layerwise energy distribution for the convolutional layers of VGG16 DNN implemented on systolic-array hardware for multi-task inference in \textit{Pipelined task mode}}
    \label{energy_pip}
     \vspace{-6mm}
\end{figure}

In Fig. \ref{energy_pip}, we plot the energy distribution of the even numbered convolutional layers of the VGG16 DNN for the \textit{Pipelined task mode}. 
We find that MIME dramatically reduces the layerwise computation and communication energies, with savings being more significant for $E_{DRAM}$ and $E_{reg}$ based memory access energies in latter convolutional layers. This is because, in the Pipelined task mode with OS dataflow, MIME greatly reduces repeated accesses to the off-chip DRAM for weights as well as thresholds. On an average, we obtain $\sim2.4-3.1\times$ savings in total energy expenditure per convolutional layer for MIME with respect to the baseline Case-1 and $\sim1.3-2.4\times$ savings with respect to the baseline Case-2. 

\begin{figure}[t]
    \centering
    \includegraphics[width=0.9\linewidth]{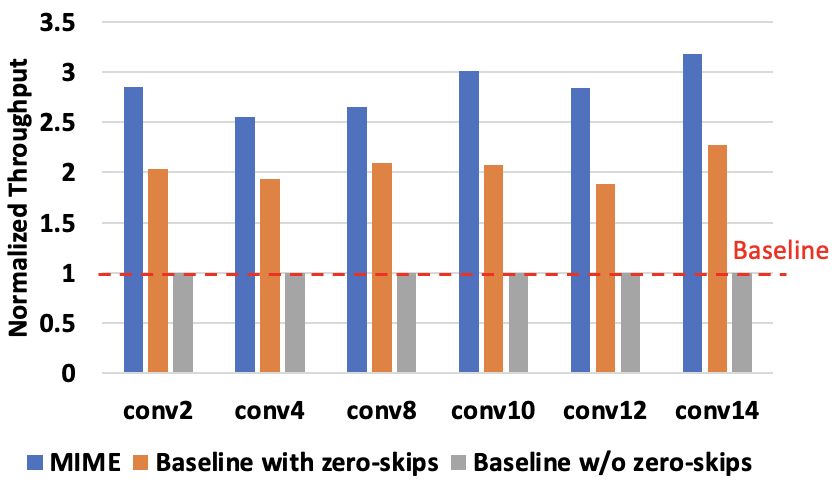}
    
    \caption{Plot showing layerwise throughput for the convolutional layers of VGG16 DNN implemented on systolic-array hardware for multi-task inference in \textit{Pipelined task mode}}
    \label{tpt_pip}
     \vspace{-2mm}
\end{figure}

In Fig. \ref{tpt_pip}, we also present a similar comparison for the improvement in throughput achieved via MIME. Here, the layerwise throughput is normalized with respect to the baseline Case-1. We find $\sim2.8-3.0\times$ improvement in throughput, that is primarily due to the reduced MAC computations in the PE arrays owing to the dynamic neuronal sparsity in MIME. 

\begin{figure}[t]
    \centering
    \includegraphics[width=0.75\linewidth]{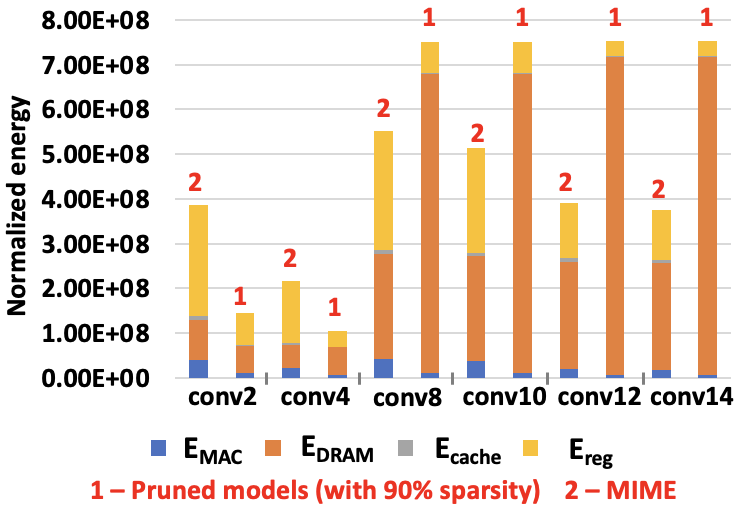}
    
    \caption{Plot showing comparison of layerwise energy expenditure in  \textit{Pipelined task mode} for MIME with respect to highly pruned models used for conventional multi-task inference}
    \label{cmp_prun}
     \vspace{-3mm}
\end{figure}

\textbf{Comparison with highly compressed/pruned models for multi-task inference:} To further evaluate the benefits of MIME in \textit{Pipelined task mode}, we compare the layerwise energy expenditure against the conventional multi-task inference approach using highly compressed/pruned ($90\%$ layerwise weight-sparsity) VGG16 models for the downstream child tasks (\textit{i.e.}, VGG16/CIFAR10, VGG16/CIFAR100 and VGG16/F-MNIST). Note, the sparse models have been generated via pruning at initialization \cite{frankle2018lottery, malach2020proving} followed by training to near iso-accuracy with the corresponding baselines in Table \ref{tab:base_acc}. The results in Fig. \ref{cmp_prun} show that for the initial convolutional layers (specifically, conv2 and conv4 layers of the VGG16 DNN), MIME under-performs with respect to inference with the pruned models. This is because in the conventional inference scenario with the pruned models, there is no requirement of fetching task-specific threshold parameters (that outnumber the weight parameters in the conv2 and conv4 layers). 
However, from conv5 layer onwards, the weight parameters outnumber the threshold parameters, and MIME prevents re-fetching the weights from the DRAM repeatedly for multiple tasks in the pipeline. Reduced DRAM accesses boosts the energy-savings in case of MIME, although its layerwise dynamic neuronal sparsity is less than $90\%$ (see Table \ref{tab:mime_acc}). The energy savings ($\sim1.36-2.0\times$) achieved by MIME in the latter convolutional layers, clearly compensates for the losses incurred in the initial conv2 and conv4 layers. 

\begin{figure}[t]
    \centering
    \includegraphics[width=0.8\linewidth]{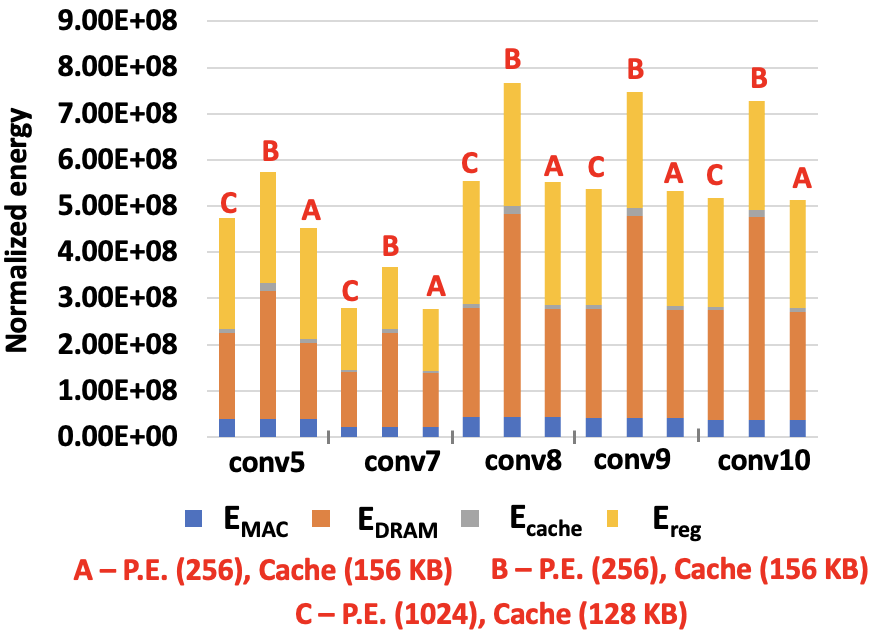}
    
    \caption{Plot to assess the impact of reducing PE array size or reducing cache memory size on layerwise energy distribution for MIME}
    \label{array_cache}
     \vspace{-6mm}
\end{figure}

\textbf{Effect of reducing cache size and PE array size:} In Fig. \ref{array_cache}, we compare three cases under MIME- \textbf{Case-A:} Usual scenario (PE array size = 1024 and Cache size = 156 KB), \textbf{Case-B:} PE array size = 256 and Cache size = 156 KB (Reduced PE array size), and \textbf{Case-C:} PE array size = 1024 and Cache size = 128 KB (Reduced cache size). We find that for the intermediate convolutional layers (conv5 to conv10), the energy expenditure increases significantly ($\sim1.26-1.41\times$) on reducing the PE array size to 256 when compared with Case-A. Specifically, this additional energy expenditure is attributed to increased DRAM accesses to fetch weight and threshold parameters to the PEs multiple times. There is also a rise in the value of $E_{cache}$, but is not as significant as $E_{DRAM}$. However, on comparing Case-C and Case-A, we find that reducing cache size does not add significant energy overhead as reducing PE array size. In summary, to extract high energy benefits from MIME in the \textit{Pipelined Task Mode}, the design should have a larger PE array over a larger cache memory to reduce repeated DRAM accesses to fetch the task-specific parameters. 

\vspace{-3mm}

\section{Conclusion}

This work proposes MIME, a technique to conduct multi-task DNN inference on a systolic-array hardware in a memory and energy-efficient manner. MIME uses the same weight parameters of a parent task ($W_{parent}$) to conduct inference for multiple child tasks. Each child task is associated with its own set of learnt threshold parameters used in conjunction with $W_{parent}$ for inference in a multi-task scenario. We show that MIME leads to significant savings in off-chip DRAM storage compared to conventional approaches to multi-task inference. An important consequence of MIME is input and task-dependent dynamic neuronal pruning that unleashes several hardware benefits. We explore a more realistic and diversified mode of inference called \textit{Pipelined task mode} and show that MIME leads to significant energy-savings and higher throughput on hardware when operated in this mode.

\section*{Acknowledgement}
This work was supported in part by C-BRIC, a JUMP center sponsored by DARPA and SRC, the National Science Foundation (Grant\#1947826) and the DARPA AI Exploration (AIE) program.

\bibliographystyle{IEEEtran}
\bibliography{reference}

\begin{thebibliography}{10}
\providecommand{\url}[1]{#1}
\csname url@samestyle\endcsname
\providecommand{\newblock}{\relax}
\providecommand{\bibinfo}[2]{#2}
\providecommand{\BIBentrySTDinterwordspacing}{\spaceskip=0pt\relax}
\providecommand{\BIBentryALTinterwordstretchfactor}{4}
\providecommand{\BIBentryALTinterwordspacing}{\spaceskip=\fontdimen2\font plus
\BIBentryALTinterwordstretchfactor\fontdimen3\font minus
  \fontdimen4\font\relax}
\providecommand{\BIBforeignlanguage}[2]{{%
\expandafter\ifx\csname l@#1\endcsname\relax
\typeout{** WARNING: IEEEtran.bst: No hyphenation pattern has been}%
\typeout{** loaded for the language `#1'. Using the pattern for}%
\typeout{** the default language instead.}%
\else
\language=\csname l@#1\endcsname
\fi
#2}}
\providecommand{\BIBdecl}{\relax}
\BIBdecl

\bibitem{krizhevsky2012imagenet}
A.~Krizhevsky \emph{et~al.}, ``Imagenet classification with deep convolutional
  neural networks,'' \emph{Advances in neural information processing systems},
  2012.

\bibitem{hinton2012deep}
G.~Hinton \emph{et~al.}, ``Deep neural networks for acoustic modeling in speech
  recognition: The shared views of four research groups,'' \emph{IEEE Signal
  processing magazine}, 2012.

\bibitem{goldberg2016primer}
Y.~Goldberg, ``A primer on neural network models for natural language
  processing,'' \emph{Journal of Artificial Intelligence Research}, 2016.

\bibitem{peng2019dnn+}
X.~Peng \emph{et~al.}, ``Dnn+ neurosim: An end-to-end benchmarking framework
  for compute-in-memory accelerators with versatile device technologies,'' in
  \emph{IEDM}, 2019.

\bibitem{chen2016eyeriss}
Y.-H. Chen \emph{et~al.}, ``Eyeriss: A spatial architecture for
  energy-efficient dataflow for convolutional neural networks,'' \emph{ACM
  SIGARCH Computer Architecture News}, 2016.

\bibitem{bhattacharjee2021neat}
A.~Bhattacharjee \emph{et~al.}, ``Neat: Non-linearity aware training for
  accurate, energy-efficient and robust implementation of neural networks on
  1t-1r crossbars,'' \emph{IEEE Transactions on Computer-Aided Design of
  Integrated Circuits and Systems}, 2021.

\bibitem{biswas2018conv}
A.~Biswas \emph{et~al.}, ``Conv-ram: An energy-efficient sram with embedded
  convolution computation for low-power cnn-based machine learning
  applications,'' in \emph{IEEE ISSCC}, 2018.

\bibitem{pan2009survey}
S.~J. Pan and Q.~Yang, ``A survey on transfer learning,'' \emph{IEEE
  Transactions on knowledge and data engineering}, 2009.

\bibitem{oquab2014learning}
M.~Oquab \emph{et~al.}, ``Learning and transferring mid-level image
  representations using convolutional neural networks,'' in \emph{IEEE CVPR},
  2014.

\bibitem{lu2021convolutional}
J.~L. Lu \emph{et~al.}, ``Convolutional autoencoder-based transfer learning for
  multi-task image inferences,'' \emph{IEEE Transactions on Emerging Topics in
  Computing}, 2021.

\bibitem{fu2021learn}
C.~Fu \emph{et~al.}, ``Learn-to-share: A hardware-friendly transfer learning
  framework exploiting computation and parameter sharing,'' in
  \emph{International Conference on Machine Learning}, 2021.

\bibitem{tan2018survey}
C.~Tan \emph{et~al.}, ``A survey on deep transfer learning,'' in
  \emph{International conference on artificial neural networks}.\hskip 1em plus
  0.5em minus 0.4em\relax Springer, 2018.

\bibitem{whatmough2019fixynn}
P.~N. Whatmough \emph{et~al.}, ``Fixynn: Efficient hardware for mobile computer
  vision via transfer learning,'' \emph{arXiv preprint arXiv:1902.11128}, 2019.

\bibitem{nurvitadhi2016hardware}
E.~Nurvitadhi \emph{et~al.}, ``Hardware accelerator for analytics of sparse
  data,'' in \emph{DATE}, 2016.

\bibitem{nurvitadhi2015sparse}
E.~Nurvitadhi, A.~Mishra, and D.~Marr, ``A sparse matrix vector multiply
  accelerator for support vector machine,'' in \emph{2015 International
  Conference on Compilers, Architecture and Synthesis for Embedded Systems
  (CASES)}, 2015.

\bibitem{yang2017designing}
T.-J. Yang \emph{et~al.}, ``Designing energy-efficient convolutional neural
  networks using energy-aware pruning,'' in \emph{Proceedings of the IEEE
  Conference on Computer Vision and Pattern Recognition}, 2017.

\bibitem{yang2018energy}
H.~Yang \emph{et~al.}, ``Energy-constrained compression for deep neural
  networks via weighted sparse projection and layer input masking,''
  \emph{arXiv preprint arXiv:1806.04321}, 2018.

\bibitem{vgg}
K.~Simonyan \emph{et~al.}, ``Very deep convolutional networks for large-scale
  image recognition,'' 2014.

\bibitem{cifar}
A.~Krizhevsky, ``Learning multiple layers of features from tiny images,'' Tech.
  Rep., 2009.

\bibitem{xiao2017fashion}
H.~Xiao \emph{et~al.}, ``Fashion-mnist: a novel image dataset for benchmarking
  machine learning algorithms,'' \emph{arXiv preprint arXiv:1708.07747}, 2017.

\bibitem{hinton2015distilling}
G.~Hinton, O.~Vinyals, and J.~Dean, ``Distilling the knowledge in a neural
  network,'' \emph{arXiv preprint arXiv:1503.02531}, 2015.

\bibitem{gou2021knowledge}
J.~Gou \emph{et~al.}, ``Knowledge distillation: A survey,'' \emph{International
  Journal of Computer Vision}, 2021.

\bibitem{liang2021pruning}
T.~Liang \emph{et~al.}, ``Pruning and quantization for deep neural network
  acceleration: A survey,'' \emph{Neurocomputing}, 2021.

\bibitem{you2020rsnn}
W.~You and C.~Wu, ``Rsnn: a software/hardware co-optimized framework for sparse
  convolutional neural networks on fpgas,'' \emph{IEEE Access}, 2020.

\bibitem{wang2020high}
J.~Wang \emph{et~al.}, ``High pe utilization cnn accelerator with channel
  fusion supporting pattern-compressed sparse neural networks,'' in \emph{DAC},
  2020.

\bibitem{sun2020computation}
F.~Sun \emph{et~al.}, ``Computation on sparse neural networks and its
  implications for future hardware,'' in \emph{2020 57th ACM/IEEE Design
  Automation Conference (DAC)}, 2020.

\bibitem{chu2020pim}
C.~Chu \emph{et~al.}, ``Pim-prune: fine-grain dcnn pruning for crossbar-based
  process-in-memory architecture,'' in \emph{2020 57th ACM/IEEE Design
  Automation Conference (DAC)}, 2020.

\bibitem{zhang2021hardware}
J.~Zhang \emph{et~al.}, ``Hardware-software codesign of weight reshaping and
  systolic array multiplexing for efficient cnns,'' in \emph{2021 Design,
  Automation \& Test in Europe Conference \& Exhibition (DATE)}, 2021.

\bibitem{parisi2019continual}
G.~I. Parisi \emph{et~al.}, ``Continual lifelong learning with neural networks:
  A review,'' \emph{Neural Networks}, 2019.

\bibitem{ramesh2021boosting}
R.~Ramesh \emph{et~al.}, ``Boosting a model zoo for multi-task and continual
  learning,'' \emph{arXiv preprint arXiv:2106.03027}, 2021.

\bibitem{liu2020dynamic}
J.~Liu \emph{et~al.}, ``Dynamic sparse training: Find efficient sparse network
  from scratch with trainable masked layers,'' \emph{arXiv preprint
  arXiv:2005.06870}, 2020.

\bibitem{frankle2018lottery}
J.~Frankle and M.~Carbin, ``The lottery ticket hypothesis: Finding sparse,
  trainable neural networks,'' \emph{arXiv preprint arXiv:1803.03635}, 2018.

\bibitem{malach2020proving}
E.~Malach \emph{et~al.}, ``Proving the lottery ticket hypothesis: Pruning is
  all you need,'' in \emph{International Conference on Machine Learning}, 2020.

\end{thebibliography}

\end{document}